\documentclass[letterpaper]{article} 
\usepackage{aaai2026}  
\usepackage{times}  
\usepackage{helvet}  
\usepackage{courier}  
\usepackage[hyphens]{url}  
\usepackage{graphicx} 
\usepackage{natbib}  
\usepackage{caption} 
\usepackage{algorithm}
\usepackage{algorithmic}
\usepackage{booktabs}
\usepackage{tabularx}
\usepackage{hyphenat}

\usepackage{newfloat}
\usepackage{listings}
\DeclareCaptionStyle{ruled}{labelfont=normalfont,labelsep=colon,strut=off} 
\floatstyle{ruled}
\newfloat{listing}{tb}{lst}{}
\floatname{listing}{Listing}
\title{Democratizing LLM Efficiency:\\From Hyperscale Optimizations to Universal Deployability}
\author {
Hen-Hsen Huang
}
\affiliations{
    Institute of Information Science, 
    Academia Sinica, 
    Taipei, Taiwan\\
    hhhuang@iis.sinica.edu.tw
}

\begin{document}
\maketitle
\begin{abstract}
Large language models (LLMs) have become indispensable, but the most celebrated efficiency methods---mixture-of-experts (MoE), speculative decoding, and complex retrieval-augmented generation (RAG)---were built for hyperscale providers with vast infrastructure and elite teams. 
Outside that context, their benefits collapse into overhead, fragility, and wasted carbon. 
The result is that a handful of Big Tech companies benefit, while thousands of hospitals, schools, governments, and enterprises are left without viable options. 
We argue that the next frontier is not greater sophistication at scale, but robust simplicity: efficiency that thrives under modest resources and minimal expertise. We propose a new research agenda: retrofitting pretrained models with more efficient architectures without retraining, inventing lightweight fine-tuning that preserves alignment, making reasoning economical despite long chains of thought, enabling dynamic knowledge management without heavy RAG pipelines, and adopting Overhead-Aware Efficiency (OAE) as a standard benchmark.
By redefining efficiency to include adoption cost, sustainability, and fairness, we can democratize LLM deployment---ensuring that optimization reduces inequality and carbon waste rather than amplifying them.
\end{abstract}


\section{From Hyperscale Myths to Everyday Reality}

\begin{table*}[t!]
    \newcommand{\leftcolsize}{2.2cm}
    \newcommand{\midcolsize}{7cm}
    \newcommand{\rightcolsize}{7cm}
    \newcommand{\linespace}{0.3cm}
    \centering
    \begin{tabular}{c c c}
        \toprule
            \noindent\parbox[c]{\leftcolsize}{\textbf{Dimension}} & 
            \noindent\parbox[c]{\midcolsize}{\textbf{Hyperscale Service Providers}} & 
            \noindent\parbox[c]{\rightcolsize}{\textbf{Small-to-Medium Service Providers}} \\ 
        \midrule
            \noindent\parbox[c]{\leftcolsize}{\nohyphens{Hardware Resources}} & 
            \noindent\parbox[c]{\midcolsize}{Massive GPU/TPU clusters; high-bandwidth interconnects; exabyte-scale storage} & 
            \noindent\parbox[c]{\rightcolsize}{Single GPUs or modest clusters; commodity hardware; limited storage capacity} \\
        \addlinespace[\linespace]
            \noindent\parbox[c]{\leftcolsize}{\nohyphens{Throughput Assumption}} & 
            \noindent\parbox[c]{\midcolsize}{Millions of requests per day; high queries-per-second; batch-friendly traffic patterns} & 
            \noindent\parbox[c]{\rightcolsize}{Low to moderate queries-per-second; sporadic or interactive requests; small batch sizes} \\
        \addlinespace[\linespace]
            \noindent\parbox[c]{\leftcolsize}{Engineering Expertise} & 
            \noindent\parbox[c]{\midcolsize}{Dedicated ML research teams; distributed systems engineers; model optimization specialists} & 
            \noindent\parbox[c]{\rightcolsize}{Small IT teams; generalist engineers; limited ML optimization expertise} \\
        \addlinespace[\linespace]
            \noindent\parbox[c]{\leftcolsize}{Optimization Focus} & 
            \noindent\parbox[c]{\midcolsize}{Pursue maximal throughput, scalability, and fine-grained efficiency gains} & 
            \noindent\parbox[c]{\rightcolsize}{Prioritize simplicity, robustness, cost-effectiveness, and ease of deployment} \\
        \addlinespace[\linespace]
            \noindent\parbox[c]{\leftcolsize}{Constraints} & 
            \noindent\parbox[c]{\midcolsize}{Few cost or scale limitations; infrastructure tailored for peak efficiency} & 
            \noindent\parbox[c]{\rightcolsize}{Limited budgets; hardware bottlenecks; strict privacy, sovereignty, or compliance requirements} \\
        \addlinespace[\linespace]
            \noindent\parbox[c]{\leftcolsize}{\nohyphens{Deployment Environment}} & 
            \noindent\parbox[c]{\midcolsize}{Global cloud services; elastic scaling; mature infrastructure support} & 
            \noindent\parbox[c]{\rightcolsize}{On-premise servers; edge or air-gapped environments; fragile infrastructure support} \\
    \bottomrule
    \end{tabular}
    \caption{Contrasting hyperscale and small-to-medium LLM service providers across resources, workloads, expertise, optimization priorities, and constraints.}
    \label{tab:providers}
\end{table*}
In theoretical computer science, asymptotic analysis teaches us to evaluate algorithms as $N \to \infty$, where constant factors are dismissed as negligible. 
Yet in practice, $N$ is usually bounded: a hospital cannot process more medical records than its patient population, which in turn is limited by the size of the city it serves. 
In such settings, the so-called ``constant'' overhead becomes decisive. 
The same misconception now pervades research on large language model (LLM) efficiency: methods optimized for hyperscale workloads may appear efficient in theory, but collapse into overhead, fragility, and wasted energy in realistic deployments.

LLMs are now widely deployed across domains, powering applications in education, customer service, law, science, and beyond. 
The most capable services today come from hyperscale cloud providers-organizations operating massive GPU clusters optimized for throughput at scale. For many mainstream use cases, these offerings are sufficient.

Yet critical and sensitive domains-such as healthcare, finance, defense, government, and highly confidential enterprises-cannot always rely on external cloud solutions. 
Concerns about sovereignty, privacy, and regulatory compliance demand in-house or on-premise deployments. 
These organizations face two persistent constraints: limited compute resources and limited in-house expertise. 
For them, the demand for efficiency is pressing but largely unmet.

Most efficiency research, however, has been shaped by hyperscale assumptions. 
Celebrated methods such as the mixture-of-experts (MOE) architectures~\citep{6797059,jordan1994hierarchical,lepikhin2020gshard,dai2024deepseekmoeultimateexpertspecialization}, speculative decoding~\citep{stern2018blockwise,xia2022speculative,leviathan2023fast,chen2023accelerating}, and complex retrieval-augmented generation (RAG) pipelines~\citep{10.1145/3637528.3671470,shen2024towards,sharma2025retrieval} were designed for environments with abundant compute and high throughput. When transplanted into small-to-medium deployments, their apparent gains often vanish, leaving only overhead and fragility.

This mismatch is not incidental but structural. 
Efficiency has increasingly been equated with ``bigger''---larger infrastructures, more intricate pipelines, and elite engineering support. Yet efficiency is contextual. 
What works in a hyperscale data center may be impractical, or even harmful, in a hospital IT department, a mid-sized business, or an air-gapped deployment. 
Today's efficiency breakthroughs serve only a handful of hyperscale providers, while thousands of hospitals, schools, governments, and enterprises remain without viable options.
True efficiency must serve the many, not the few.

This paper is directed at the research community in efficient LLMs. 
We argue that the next frontier lies not in further sophistication at hyperscale, but in addressing five grand challenges that could democratize efficiency. 
\begin{enumerate}
    \item Can pretrained LLMs be retrofitted with more efficient architectures without retraining from scratch? 
    \item Can fine-tuning become data-efficient and alignment-preserving, eliminating the need for dual model versions or costly post-alignment stages? \item How can decoding move beyond the trade-off between speed and accuracy, and make reasoning models economical despite their long chains of thought? 
    \item How do we keep LLMs up to date without the latency and fragility of heavy RAG pipelines? 
    \item Can we build an evaluation culture around Overhead-Aware Efficiency (OAE), where benchmarks reflect not only FLOPs (floating point operations per second) but also the cost of expertise, carbon, and adoption?
\end{enumerate}

Our own contributions, such as Catch-Augmented Generation (CAG)~\citep{cag} and trie-based beam search decoding~\citep{trie-based-decoding}, exemplify this philosophy: 
lightweight innovations that are model-agnostic, easy to implement, and provide immediate cost and latency savings. 
But these are only early steps. 
The broader research agenda is to rebalance efficiency so that its benefits extend beyond a handful of hyperscalers to the much larger population of organizations constrained by infrastructure, staffing, and compliance—ensuring that efficiency research reduces waste and inequality, rather than amplifying them.

\section{The Two Worlds of LLM Deployment}

The landscape of LLM deployment is sharply divided between hyperscale providers and small-to-medium service providers. 
Hyperscale organizations, such as OpenAI, Google, and Anthropic, operate massive GPU or TPU clusters connected by high-bandwidth interconnects and supported by exabyte-scale storage. 
Their workloads involve millions of queries per day, enabling them to fully exploit high-throughput optimization methods such as MoE, speculative decoding, and complex RAG pipelines. 
These providers also employ teams of specialists in machine learning (ML), distributed systems, and compiler optimization, who continually refine inference pipelines to squeeze out every possible efficiency gain. 
In such environments, even a fractional improvement in tokens per second translates to enormous cost savings at scale.

By contrast, small-to-medium service providers face a very different reality.
Hospitals, financial institutions, defense organizations, government agencies, and mid-sized enterprises often operate under strict privacy, sovereignty, or compliance constraints that prevent reliance on public cloud infrastructure. 
Their deployments are typically on single GPUs or modest clusters, often with only a few dozen gigabytes of available VRAM. 
Workloads in these environments are characterized by low to moderate throughput: sporadic requests, small batch sizes, or interactive user-facing queries. 
These organizations typically operate with generalist IT staff rather than specialized ML engineers. 
From a research perspective, this means that efficiency solutions must be designed with simplicity and robustness as first-class goals; otherwise they will remain unusable outside hyperscale labs.

This divide highlights a crucial point: methods optimized for hyperscale economics do not translate well to non-hyperscale deployments. 
Efficiency gains that depend on massive parallelism, sustained high queries-per-second (QPS), or constant engineering intervention often disappear under modest workloads, leaving only the overhead and maintenance burden behind. 
What smaller-scale providers require are efficiency techniques that emphasize robustness, simplicity, and low resource thresholds—approaches that can be deployed on commodity hardware, maintained by general IT staff, and deliver predictable improvements under real-world constraints.

Table \ref{tab:providers} summarizes these contrasts. 
Hyperscale providers optimize for throughput at scale, leveraging specialized teams and vast infrastructure, while small-to-medium providers prioritize deployability, reliability, and cost-effectiveness under limited resources. 
Recognizing this distinction is essential: without it, efficiency research risks remaining trapped in the hyperscale paradigm, leaving the vast majority of potential LLM users underserved. 
The following sections examine how over-engineered efficiency methods embody this misalignment, and why a new research agenda is needed to rebalance efficiency for broader adoption.

\section{When ``Efficiency'' Becomes Inefficiency at Small Scale}
Recent efficiency research has celebrated three approaches---MoE, speculative decoding, and complex retrieval-augmented generation (RAG). Each delivers impressive gains at hyperscale, but all share a structural weakness: their benefits collapse outside Big Tech, where workloads are modest and engineering teams are small.

\subsection{Mixture-of-Experts (MoE)}
MoE models improve efficiency at hyperscale by activating only a subset of experts per token, reducing active computation while retaining large capacity~\citep{du2022glam}.
Yet these gains rely on massive parallelism. At small scales, most experts sit idle while all must remain in memory, wasting compute and capacity~\citep{sanseviero2023moe,chitty2025moe}.
Routing overhead and synchronization costs often erase theoretical speedups, making dense models of similar active size simpler and faster to deploy~\citep{epoch2024moevsdensemodelsinference}.

\subsection{Speculative Decoding}
Speculative decoding accelerates generation by pairing a small draft model with a larger verifier~\citep{xia2022speculative,leviathan2023fast}. 
At hyperscale, the duplicated compute pays off when billions of tokens are generated daily. In modest deployments, however, the overhead of running two models, managing caches, and tuning speculation length can outweigh the benefits~\citep{su2023synergy,marzollo2024sssd}.
In many cases, standard autoregressive decoding is faster and more reliable. 

\subsection{Complex RAG Pipelines}
RAG improves factual grounding by retrieving external knowledge, but recent systems rely on multi-hop retrieval, reranking, and orchestration~\citep{lu2024turborag,zhou-etal-2025-efficiency,shen2024towards}.
These steps amortize at hyperscale but inflate time-to-first-token under sporadic queries, with retrieval latency sometimes accounting for nearly half of end-to-end delay. Beyond runtime, complex RAG multiplies dependencies: vector databases, embeddings, rerankers, and pipelines that require ongoing tuning and maintenance—burdens few non-hyperscale IT teams can manage.

\subsection{Synthesis: Complexity as Inefficiency}
Across all three cases, the pattern is consistent: scale-dependent methods mask fragility as efficiency. They assume sustained throughput, abundant parallelism, and expert engineering support. When those assumptions vanish, overhead dominates, leaving organizations with heavier pipelines but no practical gains.

The lesson for efficiency research is clear: complexity itself is a form of inefficiency. 
A method that reduces FLOPs but demands PhD-level expertise excludes the majority of adopters. 
True efficiency must be measured not only in FLOPs or tokens-per-second, but in robustness, simplicity, and ease of adoption.

\begin{table*}[t!]
    \newcommand{\firstcolsize}{2.2cm}
    \newcommand{\secondcolsize}{3.5cm}
    \newcommand{\thirdcolsize}{3.3cm}
    \newcommand{\fourthcolsize}{7cm}
    \newcommand{\linespace}{0.3cm}
    \centering
    \begin{tabular}{llll}
    \toprule
    \noindent\parbox[c]{\firstcolsize}{\nohyphens{\textbf{Topic}}} 
        & \noindent\parbox[c]{\secondcolsize}{\nohyphens{\textbf{Current Directions}}} 
        & \noindent\parbox[c]{\thirdcolsize}{\nohyphens{\textbf{Complexity Barrier}}} 
        & \noindent\parbox[c]{\fourthcolsize}{\textbf{Open Research Challenge}} \\
    \midrule
        \noindent\parbox[c]{\firstcolsize}{\nohyphens{Efficient Architectures}}
        & \noindent\parbox[c]{\secondcolsize}{FlashAttention, GQA, knowledge distillation}
        & \noindent\parbox[l]{\thirdcolsize}{\nohyphens{\textbf{Low/High}: Often drop-in, integrated into open models; extremely difficult to train from scratch}}                          
        & \noindent\parbox[c]{\fourthcolsize}{How to \textit{retrofit pretrained models} with more efficient architectures without retraining from scratch. \\ 
        Can dense models be transformed into GQA- or windowed-attention style efficiently post hoc?} \\ 
    \addlinespace[\linespace]
        \noindent\parbox[c]{\firstcolsize}{\nohyphens{Lightweight Fine-Tuning \& Alignment}}
        & \noindent\parbox[c]{\secondcolsize}{\nohyphens{LoRA, Chat Vector}}
        & \noindent\parbox[c]{\thirdcolsize}{\nohyphens{\textbf{Medium}: Requires modest ML expertise and curated data}}
        & \noindent\parbox[c]{\fourthcolsize}{How to reduce the cost of continual SFT through data-efficient selection or synthesis of ``high-value'' samples.\\
        How to preserve alignment without needing dual model versions or costly post-training.} \\
    \addlinespace[\linespace]
        \noindent\parbox[c]{\firstcolsize}{\nohyphens{Efficient Decoding}}
        & \noindent\parbox[c]{\secondcolsize}{Trie-based decoding, top-$k$ sampling, batching}
        & \noindent\parbox[c]{\thirdcolsize}{{\textbf{Medium}: Trade-offs between accuracy and latency; tuning often required}} 
        & \noindent\parbox[c]{\fourthcolsize}{How to close the gap between the speed of sampling and the accuracy of beam search. \\
        More urgently, how to make reasoning LLMs economical when chain-of-thought tokens drastically inflate decoding cost.} \\   
    \addlinespace[\linespace]
        \noindent\parbox[c]{\firstcolsize}{\nohyphens{Decoding Frameworks}} 
        & \noindent\parbox[c]{\secondcolsize}{Medusa, Skeleton-of-Thought}   
        & \noindent\parbox[c]{\thirdcolsize}{\textbf{High}: Require custom training and experimental infrastructure}       
        & \noindent\parbox[c]{\fourthcolsize}{Open research challenge: parallel or multi-branch decoding is promising but fragile. \\ 
        How can such frameworks be made robust, general-purpose, and usable without hyperscale resources?} \\ 
    \addlinespace[\linespace]
        \noindent\parbox[c]{\firstcolsize}{\nohyphens{Dynamic Knowledge Management}}
        & \noindent\parbox[c]{\secondcolsize}{\nohyphens{Prompt compression, caching, knowledge editing}}
        & \noindent\parbox[c]{\thirdcolsize}{\nohyphens{\textbf{Medium}: Practical token-saving tricks, but limited to small knowledge bases}}
        & \noindent\parbox[c]{\fourthcolsize}{The deeper issue is how to keep LLMs \textit{up to date}. \\
        How can knowledge be updated dynamically without heavy RAG pipelines? \\
        Can augmentation become an intrinsic, lightweight capability rather than an external system?} \\ 
    \addlinespace[\linespace]
        \noindent\parbox[c]{\firstcolsize}{\nohyphens{Overhead-Aware Efficiency (OAE)}}
        & \noindent\parbox[c]{\secondcolsize}{\nohyphens{Throughput-to-overhead ratio, adoption cost, robustness}}
        & \noindent\parbox[c]{\thirdcolsize}{\nohyphens{\textbf{Low}: Conceptual, but not yet standardized}}
        & \noindent\parbox[c]{\fourthcolsize}{How to rigorously quantify ``overhead,'' especially the cost of expertise. \\
        Can benchmarks incorporate the gap between elite hyperscale talent and ordinary IT engineers, making efficiency evaluations reflect real adoption barriers?} \\ 
    \bottomrule
    \end{tabular}
    \caption{Research roadmap mapping efficiency topics to their adoption barriers. The goal is not to prescribe tools for practitioners, but to highlight where future research must lower complexity and address structural barriers so that efficiency methods become usable outside hyperscale environments.}
    \label{tab:roadmap}
\end{table*}

\section{Complexity Is the New Inefficiency}
A recurring issue across Mixture-of-Experts, speculative decoding, and complex RAG pipelines is not only their scale-dependence but also their engineering complexity. These methods are typically developed and evaluated in research labs or hyperscale environments with specialized teams of systems engineers, ML researchers, and infrastructure experts. Outside of these contexts, however, such sophistication can become a liability rather than an asset.
\begin{itemize}
    \item MoE requires careful expert placement, load balancing, and communication tuning across many devices. Without deep familiarity with distributed systems and high-performance interconnects, organizations risk underutilization and instability.
	\item Speculative decoding doubles the model management problem: maintaining draft and target models, coordinating caches, tuning speculation lengths, and potentially retraining the draft model to remain aligned. This creates a fragile pipeline that demands constant monitoring and specialized knowledge to remain efficient.
	\item Complex RAG multiplies dependencies: vector databases, embedding pipelines, rerankers, index rebuilds, and orchestrating code. Each subsystem requires domain-specific expertise—retrieval tuning, database scaling, embedding fine-tuning—placing demands far beyond what a typical IT team in healthcare, finance, or government can realistically maintain.
\end{itemize}

For most organizations, IT teams are small, tasked with broad responsibilities, and cannot dedicate specialists to the constant tuning of fragile machine learning pipelines. The practical outcome is that methods celebrated as breakthroughs in academic and hyperscale contexts become impractical in deployment. Even if they deliver theoretical gains on paper, they fail the test of operational simplicity: the ability for non-expert teams to deploy, monitor, and sustain the system over time.

This raises a critical point for the efficiency research agenda: complexity itself is a form of inefficiency. 
If a method reduces FLOPs but demands PhD-level expertise to maintain, it excludes the vast majority of potential adopters. 
True efficiency must account not only for computational cost but also for the organizational cost of deployment. 
In this light, practical methods should be judged as much by their ease of adoption as by their benchmark performance.

\section{A Manifesto for Democratizing Efficiency}

If efficiency research is to move beyond hyperscale, we need a new agenda: one that values deployability, robustness, and simplicity as much as throughput benchmarks. 

Table~\ref{tab:roadmap} outlines promising directions where lightweight, broadly adoptable methods can drive real-world impact. 
Such methods help democratize LLM deployment, enabling organizations with limited ML expertise or hardware to serve advanced models cost-effectively. 

\subsection{Retrofitting Models, Not Rebuilding Them}

Most architectural efficiency gains in LLMs today come from design choices made upstream by Big Tech during pretraining. Techniques such as FlashAttention~\citep{shah2024flashattention,dao2024flashattention}, Grouped-Query Attention (GQA)~\citep{ainslie-etal-2023-gqa}, or sliding-window attention~\citep{wan2024efficient} demonstrate how much can be achieved when training is coupled with efficient design. 
Yet these innovations are effectively ``baked in'' at pretraining time, meaning non-hyperscale players---who cannot afford to retrain trillion-parameter models---must accept architectures as given.

This raises a deeper challenge: can we shift the architecture of an already pretrained model into a more efficient form without retraining from scratch? 
Instead of asking smaller organizations to wait for the next efficient model release, research could explore how to retrofit  existing open-weight models with architectural modifications post hoc. 
For example, could dense attention heads be merged or pruned into grouped structures, could windowed attention be substituted for global attention in later layers, or could memory layouts be reorganized to mimic FlashAttention-style I/O savings—all while preserving the pretrained knowledge?

Here, knowledge distillation offers an additional lever~\citep{gou2021knowledge,xu2024survey}. 
Recent advances in distilling large models into smaller ones show that knowledge can be compressed and transferred without retraining from scratch, sometimes even preserving reasoning ability or alignment~\citep{gu2024minillm}. 
Distillation techniques could therefore serve as a bridge for architectural retrofitting: rather than directly modifying a fragile pretrained backbone, a smaller distilled student could be trained to emulate the teacher's outputs while adopting more efficient architectures. 
This combination of post hoc architectural transformation and distillation-based compression points toward a practical path where efficiency gains are realized without hyperscale resources.

The technical difficulty remains substantial: both retrofitting and distillation risk accuracy degradation, loss of alignment, or domain-specific fragility. 
But if successful, these approaches would redefine efficiency as something dynamic and adaptable, not fixed at model release. 
For non-hyperscale players, this would mean efficiency is no longer dictated by Big Tech's design choices, but can evolve with deployment needs and hardware constraints.

In short, the key is not only designing efficient architectures for future models, but also developing transformation and distillation techniques that make existing models more efficient retroactively. 
Solving this challenge could democratize efficiency at a fundamental level, enabling smaller organizations to reshape models to fit their environments rather than being locked into hyperscale defaults.

\subsection{Fine-Tuning Without Fragility}

For non-hyperscale organizations, training an LLM from scratch is infeasible: the compute, data, and expertise required are far beyond the reach of most IT teams outside of Big Tech. As a result, fine-tuning open-weight models such as LLaMA, Gemma, Mistral, or Qwen has become the default strategy for local adaptation. 
While this avoids the astronomical costs of pretraining, it raises new questions about how to make supervised fine-tuning (SFT) and continual pretraining affordable and reliable in practice. 

A core challenge is reducing the cost of continual SFT. 
Full-model updates remain prohibitively expensive, and even parameter-efficient methods like LoRA~\citep{hu2021lora} require careful data curation and multiple training runs. 
How can we select or synthesize a small but highly informative set of training samples that achieve the same effect as massive fine-tuning corpora? 
Can future research design data-efficient strategies that cut costs by an order of magnitude while preserving performance?

Another persistent issue is alignment preservation. 
Many open-weight LLMs are instruction-tuned for safety and helpfulness, but these capabilities often degrade after domain-specific fine-tuning, requiring costly secondary alignment stages such as RLHF~\citep{NIPS2017_d5e2c0ad,NEURIPS2022_b1efde53,dai2024safe} or preference optimization~\citep{schulman2017proximalpolicyoptimizationalgorithms,NEURIPS2023_a85b405e,ethayarajh2024model}. 
Chat Vector~\citep{huang-etal-2024-chat} offers a clever workaround: by subtracting a base model from its instruction-tuned variant, one can extract a vector that restores alignment when added to another model. 
Yet this approach exposes deeper open problems.
First, it requires both pretrained and instruction-tuned versions of the same backbone model, while many open-weight models release only the latter. 
Second, even with Chat Vector, some performance degradation remains, meaning lightweight post-training is still necessary.

These limitations reveal broader research questions. 
Can we develop alignment-preserving fine-tuning methods that do not rely on dual versions of a backbone model? Can alignment be retained automatically, without additional reinforcement or preference optimization stages? 
Ultimately, the open challenge is to make fine-tuning data-efficient, compute-light, and alignment-stable, such that organizations with modest resources can adapt models safely without introducing new fragilities.

\subsection{Reasoning Without the Cost Explosion}
Decoding is one of the most direct levers for improving inference efficiency, yet it remains a central bottleneck. 
Conventional beam search can improve accuracy but scales linearly in latency with beam size, making it unattractive for low-resource deployments. 
Simpler alternatives such as top-$k$ or nucleus sampling require only a single forward pass per token, and are therefore faster in practice, but they often sacrifice output quality and reliability. 
Our own work on trie-based beam search~\citep{trie-based-decoding} reduces the memory and latency cost of beam search by pruning unpromising paths, but even in its most efficient form, top-$k$ sampling remains faster-albeit at the cost of accuracy.
This persistent trade-off highlights an open challenge: how to close the gap between speed and accuracy in decoding.

The challenge becomes even more urgent in the era of reasoning  LLMs~\citep{guo2025deepseek}. 
Models designed for mathematical problem solving or coding often rely on lengthy chains of thought~\citep{10.5555/3600270.3602070}. 
These ``reasoning tokens'' improve accuracy and faithfulness, but they also drastically increase decoding cost and latency~\citep{sui2025stopoverthinkingsurveyefficient}. 
A model that generates ten times more intermediate tokens may achieve better answers, yet at prohibitive runtime costs for small-scale deployments. 
This raises a pressing question: how can reasoning be performed in a more economical manner, without exploding decoding costs?

Some experimental directions—such as Medusa~\citep{10.5555/3692070.3692273} and Skeleton-of-Thought~\citep{ning2024skeletonofthought}—attempt to parallelize the generation of reasoning paths, but they require custom model training and remain fragile outside research settings.  
Serving-level optimizations like batching in vLLM~\citep{kwon2023efficient} improve throughput, yet they do not fundamentally reduce the cost of long reasoning traces. 
The open problem is deeper: can we design decoding strategies that preserve the reliability of reasoning while constraining its token and latency footprint?

This challenge reframes decoding not as an engineering detail but as a frontier research problem. 
To democratize reasoning LLMs, we need methods that make inference both accurate and efficient, ensuring that long-form reasoning does not remain a luxury affordable only to hyperscalers.

\subsection{Knowledge That Updates Itself}
RAG has become the standard solution for grounding LLMs, but its growing complexity-multi-hop retrieval, cross-encoder reranking, orchestration layers-comes at the cost of latency and fragility. 
CAG~\citep{cag} offers a simpler alternative by preloading reusable contexts, but it cannot scale to scenarios where knowledge sources are large. 
Even the simplest RAG pipeline inevitably introduces retrieval latency into response time, meaning that efficiency bottlenecks remain unsolved as long as we treat augmentation as an external pipeline.

This raises a deeper research problem: how should we keep LLMs up to date, and how can we manipulate their knowledge dynamically without incurring heavy retrieval overhead? 
Current RAG-based techniques like prompt compression, redundancy elimination, and caching highlight practical steps toward reducing token load and latency, but they do not address the fundamental limitation: the model's knowledge is static at release, forcing us to bolt on retrieval systems that reintroduce cost and complexity. Attempts at knowledge editing~\citep{meng2022locating}, which lightly revise a subset of model weights to inject or correct knowledge, offer a more intrinsic alternative. 
However, even the most recent advances such as AlphaEdit~\citep{fang2025alphaedit} remain far from practical deployment, highlighting the need for research that makes knowledge updates efficient, robust, and broadly usable.

Future research must therefore move beyond ``pipeline efficiency'' toward knowledge dynamism as an architectural property. 
Can we design LLMs that integrate retrieval seamlessly into the decoding process, so that grounding does not inflate latency? 
Can we update or swap knowledge modules post-training without retraining the entire model? 
Can models learn to manage their own context economically, deciding when to retrieve, when to cache, and when to rely on pretrained memory?

The challenge is not simply to compress tokens or rerank faster, but to redefine augmentation as a lightweight, intrinsic capability of LLMs rather than a fragile external system. 
Solving this problem would make LLMs both current and efficient, ensuring that non-hyperscale players can deploy models that stay relevant without incurring prohibitive retrieval overheads. 

\subsection{Measuring What Matters: Overhead, Fairness, and Carbon}

Current efficiency benchmarks focus narrowly on FLOPs, latency, or tokens-per-second. 
While these metrics capture throughput at hyperscale, they fail to measure the hidden costs that dominate outside Big Tech.
For small-to-medium deployments, the central bottleneck is not FLOPs but overhead---the additional human and organizational cost of adopting and sustaining a method.

We propose the concept of OAE: a framework in which efficiency is measured not only in raw computational terms but also in the cost of expertise required. 
A method that looks efficient on paper may demand months of work by elite ML engineers to deploy reliably. 
Hyperscale companies can absorb these costs, but most organizations rely on small IT teams with limited ML expertise.
For them, the same ``efficient'' method may yield no net benefit, or worse, introduce new fragilities and energy waste.

The open challenge is how to quantify overhead rigorously. 
Should we measure adoption cost in engineer-weeks? 
Should we model the monetary gap between hyperscale ML specialists and ordinary IT engineers? 
Can we define robustness benchmarks that reflect messy, real-world traffic where fine-tuning or retraining may not be feasible? 
These questions point to the need for new evaluation practices that bridge the talent asymmetry between Big Tech and everyone else.

In addition to traditional metrics, OAE emphasizes:
\begin{itemize}
    \item \textbf{Adoption cost}: how many engineer-weeks, and at what level of expertise, are needed to deploy and sustain a method.
    \item \textbf{Robustness under constraint}: whether the method continues to work under noisy inputs, irregular traffic, or commodity hardware.
    \item \textbf{Talent dependence}: whether efficiency gains require hyperscale-level expertise, or can be maintained by generalist IT teams.
\end{itemize}

Ultimately, OAE reframes efficiency as a property of both algorithms and organizations. 
In this broader view, the most valuable techniques are not those that squeeze out an extra 5\% tokens/sec at hyperscale, but those that minimize expertise gaps and sustain favorable throughput-to-overhead trade-offs across scales. 
By adopting OAE, the community can ensure that efficiency innovations are not reserved for a few hyperscalers but become accessible to the many.

\section{Efficiency That Serves the Many, Not the Few}
The trajectory of LLM efficiency research has been shaped by hyperscale assumptions: massive clusters, elite engineering teams, and workloads large enough to amortize fragile pipelines. 
Methods such as MoE architectures, speculative decoding, and multi-stage RAG pipelines impressive results under these conditions, but for the vast majority of organizations they collapse into overhead and fragility. 
When non-hyperscale adopters attempt to implement them, they may not only fail to see benefits but also introduce inefficiencies that waste energy and carbon resources. 
The outcome is a widening competition gap: a few hyperscale providers reap the rewards, while thousands of hospitals, schools, governments, and enterprises are left behind. 

We argue for a different vision: efficiency must be redefined as overhead-aware, fairness-oriented, and sustainable. 
Our research agenda highlights five grand challenges: Can pretrained LLMs be retrofitted into more efficient architectures without retraining from scratch? Can fine-tuning become data-efficient and alignment-preserving? Can decoding be redesigned so that reasoning models remain economical despite long chains of thought? Can augmentation evolve into dynamic knowledge management, keeping LLMs up to date without the cost of heavy retrieval? And finally, can we institutionalize OAE, building benchmarks that capture not only FLOPs but also expertise, adoption cost, and environmental impact?

True efficiency must be measured not only in tokens per second, but in who can use it, how much energy it consumes, and whether it narrows or widens inequality. By pursuing robust, simple, and sustainable methods, the research community has the opportunity to ensure that optimization serves the many, not just the few---and to build a more equitable, energy-conscious future for AI.

\section{Acknowledgments}
This research was partially supported by the National Science and Technology Council (NSTC), Taiwan, under Grant No. 112-2221-E-001-016-MY3, and by Academia Sinica under Grant No. 236d-1120205.

\bibliography{aaai2026}

\end{document}